\pdfoutput=1

\documentclass[11pt]{article}
\usepackage{acl}
\usepackage{times}
\usepackage{url}
\usepackage{latexsym}
\usepackage{tikz}
\usepackage{booktabs}
\usepackage{inconsolata}
\usepackage{amsmath}
\usepackage{float}
\usepackage{amsfonts}
\usepackage[framemethod=tikz]{mdframed}
\usetikzlibrary{positioning, shapes.geometric, arrows.meta, calc}

\definecolor{mycolor}{rgb}{0.122, 0.435, 0.698}

\newmdenv[innerlinewidth=0.5pt, roundcorner=4pt,linecolor=mycolor,innerleftmargin=6pt,
innerrightmargin=6pt,innertopmargin=6pt,innerbottommargin=6pt]{mybox}

\title{Long Context Automated Essay Scoring with Language Models}

\author{Christopher Ormerod \\
  Cambium Assessment Inc. \\ 
  {\tt christopher.ormerod@cambiumassessment.com} \\\\
  Gitit Kehat \\
  Cambium Assessment Inc. \\
  {\tt gitit.kehat@cambiumassessment.com} \\}

\date{}

\begin{document}
\maketitle
\begin{abstract}
Transformer-based language models are architecturally constrained to process text of a fixed maximum length. Essays written by higher-grade students frequently exceed the maximum allowed length for many popular open-source models. A common approach to addressing this issue when using these models for Automated Essay Scoring is to truncate the input text. This raises serious validity concerns as it undermines the model's ability to fully capture and evaluate organizational elements of the scoring rubric, which requires long contexts to assess. In this study, we evaluate several models that incorporate architectural modifications of the standard transformer architecture to overcome these length limitations using the Kaggle ASAP 2.0 dataset. The models considered in this study include fine-tuned versions of XLNet, Longformer, ModernBERT, Mamba, and Llama models.
\end{abstract}

\section{Introduction}

Automated Essay Scoring (AES) is the application of statistical models to approximate the grading of essays by a human using a rubric. The initial models employed for AES were based on word frequencies and hand-crafted features \cite{page_project_2003}. The methods and models applied to AES have closely followed those used in more general Natural Language Processing (NLP) applications. The models employed in AES include recurrent and convolutional neural networks \cite{taghipour_neural_2016}, models with attention mechanisms \cite{dong_attention-based_2017}, and transformer-based large language models (LLM) \cite{rodriguez_language_2019}. Currently, LLMs are readily used to perform AES in research and large-scale assessment \cite{lottridge_psychometric_2023}.  

The first transformer-based LLM to be applied to AES was the Bidirectional Encoder-based Representations by Transformers (BERT) \cite{devlin_bert_2018}. Since BERT arrived on the scene, the BERT model and its derivatives have readily provided state-of-the-art results in a wide range of downstream NLP tasks \cite{wang_glue_2019}. The key to the success of these LLMs has been due to the transformer architecture \cite{vaswani_attention_2017} and to the ability to pretrain the model weights on a large corpus of unlabeled data on a semisupervised task such as next-token prediction \cite{radford_improving_2018} or masked-word prediction \cite{devlin_bert_2018}. While we often say that the pretraining provides the model with some limited ``understanding", the model weights are simply encoding enough information to encode the necessary word-probability functions. 

Transformer-based models are deep feed-forward networks utilizing residual connections between layers that help stabilize training and prevent vanishing gradients \cite{vaswani_attention_2017}. Each layer uses a multiheaded attention mechanism, similar to those used in recurrent networks \cite{graves_hybrid_2013}. The input is defined by the addition of a positional embedding and a word embedding, which also defines the fixed length of the feedforward network. Since the computing power required by the attention mechanism scales quadratically with length, the length chosen for BERT was 512 \cite{devlin_bert_2018}. This length became something of a standard for the most popular transformer-based LLMs. 

The need for models that could overcome the limitations imposed by the transformer architecture became an active area of research shortly after BERT's release. We selected five different models that employ distinct approaches to addressing this challenge. These include versions of XLNet \cite{yang_xlnet_2019}, Longformer \cite{beltagy_longformer_2020}, ModernBERT \cite{warner_smarter_2024}, Mamba \cite{gu_mamba_2024}, and a generative Llama model \cite{aimeta_llama_2024} fine tuned for scoring using parameter-efficient methods \cite{xu_parameter-efficient_2023}. We give a brief explanation as to how each of these models addresses this limitation in \S \ref{sec:models}. The most novel of these approaches is applied in the Mamba model, which is the only pretrained language model in this study that uses the state-space model (SSM) \cite{gu_efficiently_2021}. For SSMs, the computing power required scales linearly with the length of the input.

To understand the limitations of AES, researchers introduced the Automated Student Assessment Prize (ASAP) Dataset using the Kaggle platform \cite{shermis_contrasting_2013}. This Dataset consists of essay responses to eight prompts, some of which were assessed using trait scoring and some of which were assessed using a simpler holistic rubric. While this dataset became the definitive benchmark for AES methods, most essay responses possessed fewer than 512 tokens. This meant that, while LLMs showed superior performance with respect to traditional AES criteria \cite{williamson_framework_2012}, the dataset did not adequately test the length issues that are often critical in the application of LLMs in large-scale assessment \cite{lottridge_psychometric_2023}. 

A second dataset, known as the Persuasive Essays for Rating, Selecting, and Understanding Argumentative and Discourse Elements (PERSUADE) corpus \cite{crossley_persuasive_2022}, which was originally designed to evaluate the performance of models that annotate the argumentative components of essays, was later extended to the Automated Student Assessment Prize v2 (ASAP 2.0) \cite{crossley_large-scale_2025}. We will describe the dataset in more detail below, but many responses in the ASAP 2.0 dataset are too long for most language models. 

This article is organized as follows: We use \S \ref{sec:models} to highlight the characteristically different approaches of the models chosen for this study. This is followed by \S \ref{sec:methods} in which we describe the data used and the training methods. We have two different training regimes: one regime for classification models, such as those obtained by appending a classification, and another regime for generative LLMs. This is followed by the results in \S \ref{sec:results} and a discussion in \S \ref{sec:discussion}.

\section{Models}\label{sec:models}

In this section, we discuss each model used in this study and why we chose to include it. We have attempted to illustrate if and how these models circumvent the architecturally imposed length restrictions of the standard transformer architecture.

\subsection{DeBERTa}

The DeBERTa model has a context length of 512. It has been chosen for this study to provide a strong benchmark for models typically used for AES. It is widely regarded as one of the best-performing models in a range of tasks. The model was trained as a discriminator, similarly to the ELECTRA models \cite{clark_electra_2020}. The DeBERTa models also deviate from the standard BERT model by disentangling the word-embedding from the positional embedding \cite{he_debertav3_2021}.

\subsection{Longformer}

The Longformer model attempts to reconcile the need for local attention with a selective form of global attention. The local attention is applied in the form of a sliding window, similar to attention using convolutional units \cite{wu_pay_2019} coupled with a form of global attention only applied to special tokens \cite{beltagy_longformer_2020}, such as the beginning, ending, and mask tokens. This model still possesses a length limitation, however, by only using attention selectively, the computational burden is mitigated, allowing for pretraining over larger context lengths. 

\subsection{XLNet}

The XLNet model uses the recurrent definition of attention introduced by the Transformer-XL model \cite{dai_transformer-xl_2019}. These models have recently been discussed for essays, where the long context was useful in accurately annotating the argumentative components of essays \cite{ormerod_argumentation_2023}. Almost all masked-language models are encoder-only models; however, the XLNet model is also distinguished as one of the few decoder models that was autoregressively pretrained as a masked-language model \cite{yang_xlnet_2019}. 

To demonstrate the recurrence, suppose any input sequence of length $L$ is denoted $s_\tau = [x_{\tau,1}, \ldots, x_{\tau,L}]$ while the hidden state for $n$-th layer associated with $s_{\tau}$ is $h_{\tau}^n \in \mathbb{R}^{L\times d}$. The recurrence relation defining $h_{\tau+1}^n$ as a function of $h_{\tau}^{n-1}$ and $h_{\tau+1}^{n-1}$ is given as follows:
\begin{subequations}\label{eq:recurrence}
\begin{eqnarray}
\tilde{h}_{\tau+1}^{n-1} &=& [SG(h_{\tau}^{n-1}) \circ h_{\tau+1}^{n-1}], \\
q_{\tau+1}^{n} &=& h_{\tau+1}^{n-1} W_q,\\ 
k_{\tau+1}^{n} &=& \tilde{h}_{\tau+1}^{n-1} W_k,\\
v_{\tau+1}^{n} &=& \tilde{h}_{\tau+1}^{n-1} W_v,\\
h_{\tau+1}^{n} &=& \mathrm{MHA}(q_{\tau+1}^{n}, k_{\tau+1}^{n}, v_{\tau+1}^{n}),
\end{eqnarray}
\end{subequations}
\noindent where $SG$ is the stop gradient, $[x\circ y]$ is the concatenation operation of two sequences, and $\mathrm{MHA}$ is an abbreviation for the typical multiheaded attention mechanism for the transformer layer. The recurrence is built into the definition of $\tilde{h}_{\tau}^{n}$, affecting the keys and values. Digging deeper into \eqref{eq:recurrence} tells us that while the definition allows for infinite input lengths, there is a functional limitation of the architecture in which the output of any token is only a function of at most $LD$ of the previous tokens where $D$ is the depth of the network. The base and large pretrained models released with \cite{yang_xlnet_2019} has $L=512$ and $D = 12$ and $D = 24$ respectively. This effectively caps the practical length to $6,000$ and $12,000$ for these models, respectively.

\subsection{ModernBERT}

The ModernBERT model is an encoder-based masked language model benefiting from much of the research that has been conducted since BERTs release \cite{warner_smarter_2024}. In particular, applications of generative LLMs have pushed the context length limitations in ways that the previous models stated above have not. The key to the context length of 8196 has been the Rotational Position Embedding (RoPE) \cite{su_roformer_2024}. There is a pretraining step in which the model is trained at short lengths with a large rotational component, then further trained on a model that interleaves rotational embedding with small and large rotational values to capture contributions from close and distant tokens. This method, developed in \cite{fu_data_2024}, was key to extending the context length for a range of popular models such as the herd of Llama models \cite{aimeta_llama_2024}. 

\subsection{Llama}

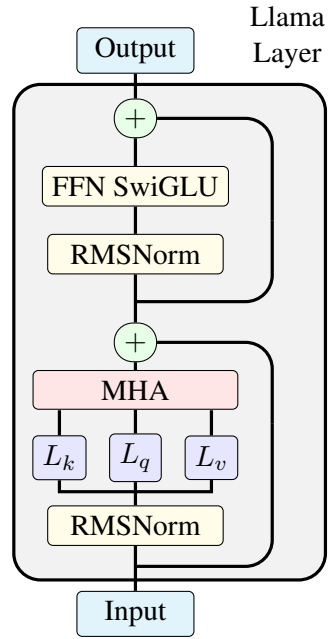
\begin{figure}[!ht]
\begin{center}
\begin{tikzpicture}[yscale=0.9]
\draw[fill= black!5, very thick,rounded corners=10pt] (-.6,-3.3) rectangle (3.5,4);


\node[rounded corners=2pt, draw=black, fill=yellow!10, minimum width=6em](rms2) at (1,-2.5) {RMSNorm};

\node[rounded corners=2pt, draw=black, fill=red!10, minimum width=7em](mha) at (1,-.5) {MHA};
\node[rounded corners=2pt, draw=black, fill=blue!10, minimum width=1em](k) at (0,-1.5) {$L_k$};
\node[rounded corners=2pt, draw=black, fill=blue!10, minimum width=1em](q) at (1,-1.5) {$L_q$};
\node[rounded corners=2pt, draw=black, fill=blue!10, minimum width=1em](v) at (2,-1.5) {$L_v$};

\node[rounded corners=2pt, draw=black, fill=yellow!10, minimum width=6em](ffn) at (1,2.5) {FFN SwiGLU};
\node[rounded corners=2pt, draw=black, fill=yellow!10, minimum width=6em](rms1) at (1,1.5) {RMSNorm};
\node[rounded corners=7pt, draw=black, fill=green!10](add1) at (1,0.2) {$+$};
\node[rounded corners=7pt, draw=black, fill=green!10](add2) at (1,3.5) {$+$};

\node[rounded corners=2pt, draw=black, fill=cyan!10, minimum width=4em](in) at (1,-3.8) {Input};
\node[rounded corners=2pt, draw=black, fill=cyan!10, minimum width=4em](out) at (1,4.5) {Output};

\draw[very thick] (in) -- (rms2);
\draw[very thick] (rms2)--(q);
\draw[very thick] (q) -- (mha);
\draw[very thick] (mha)--(add1);

\draw[very thick] (k) -- (0,-.8);
\draw[very thick] (v) -- (2,-.8);
\draw[very thick] (1,-2) -| (v);
\draw[very thick] (1,-2) -| (k);

\draw[very thick] (add1) -- (rms1);
\draw[very thick] (rms1) -- (ffn);
\draw[very thick] (add2) -- (ffn);
\draw[very thick] (add2) -- (out);

\draw[very thick, rounded corners=10pt] (1,-3.1) -| (2.8,-1);
\draw[very thick, rounded corners=10pt] (2.8,-1) |- (add1);
\draw[very thick, rounded corners=10pt] (1,.8) -| (2.8,2);
\draw[very thick, rounded corners=10pt] (2.8,2) |- (add2);
\node at (3,4.7) {\begin{tabular}{c} Llama \\ Layer \end{tabular}};
\end{tikzpicture}
\end{center}
\caption{A layer of the Llama decoder-only architecture.\label{fig:llama}}
\end{figure}

The Llama series is a family of open-source generative LLMs from Meta  \cite{aimeta_llama_2024}. The models have become as ubiquitously associated with open-source generative models as BERT was to masked language models. These generative models use RoPE \cite{su_roformer_2024} in combination with the methods used to extend context lengths to 128k \cite{fu_data_2024}. In terms of architecture, the Llama models are a variant of the decoder-only transformer-based models, utilizing RMSNorm layers and a particular activated fully connected layer. We present this architecture in Figure \ref{fig:llama}, paying particular attention to the linear layers normalizing the input into the multi-headed attention (MHA) mechanism. 

As a generative model, it was trained to predict the next token \cite{radford_improving_2018}, followed by instruction tuning \cite{chung_scaling_2022}, followed by a reinforcement learning phase to make the models more useful \cite{kaufmann_survey_2024}. These models come in a variety of sizes. The latest models include multi-modal capabilities; however, the models employed in this article are limited to text. 

\subsection{State-Space Models}

\begin{figure}[!ht]
    \centering
    \begin{tikzpicture}[yscale=0.9]
    \node[rounded corners=2pt, draw=black, fill=cyan!10, minimum width=4em](in) at (0,-4.3) {Input};
\node[rounded corners=2pt, draw=black, fill=cyan!10, minimum width=4em](out) at (0,3.3) {Output};

    \draw[fill= black!5, very thick,rounded corners=10pt] (-1,-3.7) rectangle (3,2.7);
    \node[rounded corners=2pt, draw=black, fill=red!10](ssm) at (0,0) {SSM};
    \node[rounded corners=7pt, draw=black, fill=green!10](s1) at (0,-1) {$\sigma$};
    \node[rounded corners=7pt, draw=black, fill=green!10](s2) at (2,-.5) {$\sigma$};
    \node[rounded corners=7pt, draw=black, fill=green!10](t) at (0,1) {$\times$};
    
    \node[rounded corners=2pt, draw=black, fill=yellow!10,minimum width=3em](c1) at (0,-2) {conv};
    \node[rounded corners=2pt, draw=black, fill=blue!10,minimum width=4em](in1) at (0,-3) {$L_{in}$};
    \node[rounded corners=2pt, draw=black, fill=blue!10,minimum width=4em](in2) at (2,-1.5) {$L_{gate}$};
    \node[rounded corners=2pt, draw=black, fill=blue!10,minimum width=4em](out) at (0,2) {$L_{out}$};
    \draw[very thick](in1)--(c1);
    \draw[very thick](c1)--(s1);
    \draw[very thick](s1)--(ssm);
    \draw[very thick](ssm)--(t);
    \draw[very thick](t)--(out);
    \draw[very thick](s2)|-(t);
    \draw[very thick](in2)--(s2);
    \draw[very thick] (0,-4) -- (in1);
    \draw[very thick] (0,-3.5) -| (in2);
    \draw[very thick] (out) -- (0,3);
    \node at (2,3.3) {\begin{tabular}{c} Mamba \\ Layer \end{tabular}};
    \end{tikzpicture}
    \caption{A single layer of the Mamba model. \label{fig:mamba}}
\end{figure}
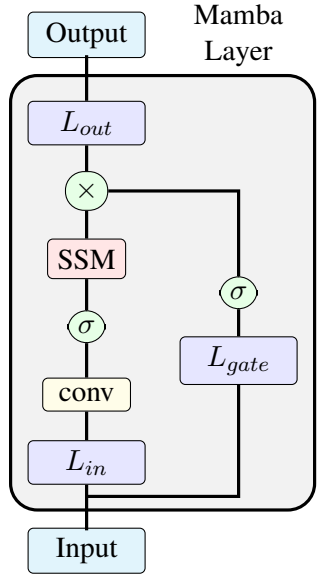

This novel architecture completely replaces the transformer layer and attention with a simpler system based on discretizations of the state-space model (SSM). The SSM is a family of differential equations specified by the matrix equations  
\begin{subequations}\label{eq:ssm}
\begin{eqnarray}
x'(t) &=& Ax(t) + Bu(t),\\
y(t) &=& Cx(t) + Du(t),
\end{eqnarray}
\end{subequations}
where $x$, $u$, and $y$ are vectors and $A,B,C$, and $D$ are matrices. This is a class of models broadly used in control theory. A standard discretization of \eqref{eq:ssm} provides us with the recurrence relation of the form
\begin{subequations}\label{eq:dssm}
    \begin{eqnarray}
h_t &=&  A h_{t-1} + B x_t,\\
x &=& C h_t.
\end{eqnarray}
\end{subequations}
A Mamba Layer, in contrast with the Transformer Layer, uses \eqref{eq:dssm} as one component in addition to linear projections, a convolutional layer, and activation functions, as shown in Figure \ref{fig:mamba}.

The Mamba blocks can be computed with linear complexity, making them well-suited for long context tasks \cite{gu_efficiently_2021}. This claim has been validated empirically by the superior performance of the Jamba models, which is an ensemble of transformer and Mamba layers \cite{lieber_jamba_2024}, on RULER benchmarks \cite{hsieh_ruler_2024}. As we seek longer and longer context lengths, models with linear complexity may be favorable from an efficiency standpoint. 

\subsection{Data}

The reason we chose the ASAP 2.0 dataset \cite{crossley_large-scale_2025} is that this dataset provides a much-needed update of the original ASAP dataset \cite{shermis_contrasting_2013}, which could be considered to be saturated at this point. This dataset, derived as an extension of the PERSUADE corpus \cite{crossley_persuasive_2022}, consists of essays written by students from grades 6 to 10 on a wide range of prompts. 

Since a key feature of this study is our ability to handle long contexts, it is important to consider the length and grade level characteristics of the data. Because we are using a variety of LLMs, each of which has adopted different subword tokenizations \cite{kudo_sentencepiece_2018}, we have no unified notion of what defines a token. In lieu of a uniform tokenization, we will report the word count reported in the dataset. These length characteristics have been presented in Table \ref{tab:length}.

\begin{table}[!ht]
    \centering
    \begin{tabular}{l|r r | r r } \toprule
         & \multicolumn{2}{c |}{Train} & \multicolumn{2}{c}{Test} \\ \midrule
         &        & Avg. & & Avg. \\
   Grade &  Count & Words & Count & Words \\ \midrule
     6  & 2094 & 292.2 & 527 & 268.3\\
     8  & 1648 & 339.9 & 921 & 295.9\\
     9  & 4002 & 426.1 & 0 & -\\ 
     10 & 9563 & 385.8 & 5973 & 356.4\\ \midrule
    Total & 17307&376.1 & 7421 & 342.7\\ \bottomrule
    \end{tabular}
    \caption{The size and length characteristics of the ASAP 2.0 dataset.}
    \label{tab:length}
\end{table}

To evaluate the data, we use the standard metrics specified for AES \cite{williamson_framework_2012}. The main metric used is the agreement statistic known as quadratic weighted kappa (QWK). Generally, the weighted kappa is specified by the equation 
\begin{equation}\label{eq:qwk}
\kappa = 1 - \dfrac{\sum_{i,j} W_{i,j} O_{i,j}}{\sum_{i,j} W_{i,j} E_{i,j} }
\end{equation}
where $O_{i,j}$ is the observed agreement between the first rater giving a score of $i$ and the second rater a score of $j$, and $E_{i,j}$ is the expected agreement only assuming the two raters' general distribution. This becomes QWK under the weighting 
\[
W_{i,j} = \frac{(i-j)^2}{(n-1)^2}
\]
where $n$ is the number of scores. It is generally understood that this is a measure of agreement above random chance, where a QWK of 1 is perfect agreement and -1 is perfect disagreement. In practical terms, lower scores represent the level of reliability between raters \cite{mchugh_interrater_2012}, and our models should be compared against human-human agreements \cite{williamson_framework_2012}. The QWK between the raters is reported to be $0.745$

\section{Methods}\label{sec:methods}

In order to perform essay scoring using LLMs, we distinguish two different cases. We call the first case traditional LLM-based scoring, where the underlying LLM is a masked-language model, such as BERT \cite{devlin_bert_2018}, or a next word predictor such as the Generative Pretrained Transformer (GPT) \cite{radford_improving_2018}. The second class of models considered was generative, which are distinguished by typically possessing an order of magnitude more parameters, and being trained in three phases: pretaining, instruction tuning, and reinforcement \cite{openai_gpt-4_2023}. 

\subsection{Traditional LLM based scoring}

The typical procedure for traditional scoring is to convert a next word or masked word prediction model into a classifier by removing the linear head that would otherwise predict a token and append, in its place, a classification head with as many targets as there are scores \cite{rodriguez_language_2019}. The classification head is randomly initialized. 

To train each of these models, 10\% of the training set was designated as a development set. The models were trained by applying the Adam optimizer with a weight decay mechanism \cite{loshchilov_decoupled_2019} to the cross-entropy loss function. An initial learning rate of $10^{-6}$ and a linear learning rate scheduler that reduces the learning rate to 0 over 10 epochs was used with a batch size of either 4 or 1 due to the length of some essays. The QWK was optimized on the development set using an early stopping mechanism. 

To fine-tune our Mamba models for classification, we appended a learnable classification head, however, we were required to effectively freeze the weights associated with the SSM, $L_{gate}$, and the convolutional layer (See Figure \ref{fig:mamba}). Full model training seemed to readily lead to model collapse, perhaps due to the requirement that certain weights take a particular form \cite{gu_efficiently_2021}. Hence, we fine-tuned the embedding layer and the associated $L_{in}$ and $L_{out}$ weights of every layer. This is a memory-efficient way to fine-tune that provides excellent results. We used the Adam optimizer above with a learning rate of $10^{-5}$ and a batch size of 8.

\subsection{Generative LLM based scoring}

Many attempts in the literature seek to optimize the prompting of closed-source generative models to yield higher agreement rates \cite{xiao_automation_2024}. While this is an interesting approach, we believe fine-tuning is necessary to obtain reasonable success. Due to the large size of the models, in order to do this with reasonable computational resources, we need to employ parameter-efficient methods \cite{xu_parameter-efficient_2023}. These methods can be applied without reference to an API and, hence, can be effectively employed securely, and privately, generating a fraction of the carbon emissions \cite{bulut_rise_2024}.

In the case of fine-tuning generative models, the dataset used mimics an instruction set the model has been trained on. This means that any element of the training set appears to be a user prompting the model to score an essay to a rubric \cite{ormerod_automated_2024}. To do this, we used the following prompt template:

\begin{mybox}
{\bf User}
\begin{verbatim}
Assign a **Score** to the 
**Essay** using the **Rubric** 
provided. 

**Rubric**: {rubric}             

**Essay**:
\end{verbatim}
\end{mybox}
\begin{mybox}
{\bf Assistant}
\begin{verbatim}
**Score**: {score}
\end{verbatim}
\end{mybox}

This template highlights the important aspects by using markdown, due to the formatting of the corpus the model was trained on. Given that variations in prompting can have a significant bearing on the results, we exploit this by allowing the model to summarize and rephrase the rubric in 20 different ways. We optimized the variation of the rubric by evaluating the QWK of the model before fine-tuning on a development set that consisted of 10\% of the training set. 

We apply the method of low-rank adapters \cite{hu_lora_2021} and quantization (QLoRA) by \cite{dettmers_qlora_2023}. To apply QLoRA to a model, we must specify which linear layers to apply the adapter to, the rank of the adapter, scaling factors, the usual learning rate, and batch size. Concerning Figure \ref{fig:llama}, we seek to apply low-rank adapters to $L_q$, $L_k$, and $L_v$ in the Llama model.

\section{Results}\label{sec:results}

The study evaluated various long-context language models on the ASAP 2.0 dataset to assess their effectiveness in automated essay scoring (AES). Models tested included traditional encoder-only architectures like DeBERTa-Base and XLNet-Base, extended-context models such as Longformer and ModernBERT, a state-space model (Mamba-130m), and generative decoder-based models like Llama-3.2-8B.

\begin{table*}
\centering
\begin{tabular}{l|l r r | r | r  r r} \toprule
&&&& \multicolumn{4}{c}{Grade} \\
Model & Reference & L & Model Size & Overall & 6 & 8 & 10 \\ \midrule
Human & \cite{crossley_large-scale_2025} & $\inf$ & & 0.745 \\
DeBERTa-Base & \cite{he_debertav3_2021} & 512 & 183M & 0.790 & 0.696 & 0.659 & 0.800 \\
XLNet-Base  & \cite{yang_xlnet_2019} & 8k${}^{*}$ & 110M & 0.784 & 0.654 & 0.640 & 0.798\\
Longformer & \cite{beltagy_longformer_2020} & 4k& 149M & 0.798 & 0.698 & 0.658 & 0.811 \\
ModernBERT & \cite{warner_smarter_2024} & 8k & 149M  & 0.790 & 0.639 & 0.658 & 0.804\\
Mamba-130m & \cite{gu_mamba_2024} & 8k${}^{*}$ & 130M & 0.797 & 0.674 & 0.640 & 0.812\\
Llama-3.2-8B & \cite{aimeta_llama_2024} & 8k & 8B & 0.792 & 0.667 & 0.672 & 0.803 \\ \bottomrule
\end{tabular}
\caption{The performance of each model in terms of QWK, given by \eqref{eq:qwk}. These context lengths for XLNet models and Mamba models are not specified. The value of 8k was implemented as a mechanism to bound the memory required for training.}
\end{table*}

Human-human rater agreement stood at 0.745, serving as the baseline for comparison. All models surpassed this baseline, with Longformer achieving the highest overall QWK of 0.798. Notably, Mamba-130m performed competitively despite its smaller parameter size, demonstrating that linear-complexity models can rival attention-based transformers in AES tasks. Key findings revealed that long-context models, particularly those using advanced architectural innovations like RoPE-based positional embeddings and selective state spaces, are well-suited for handling lengthy student essays. Traditional models like DeBERTa and XLNet showed strong performance but lagged slightly behind Longformer and Mamba. Despite their large parameter counts and sophisticated training methods -- such as instruction tuning and reinforcement learning —- generative models did not significantly outperform encoder-based models. However, they do offer the promising capability of providing feedback \cite{ormerod_automated_2024}.

\section{Discussion}\label{sec:discussion}

Overall, the results affirm the viability of long-context models in automated scoring systems, especially when dealing with complex, lengthy texts where global coherence and argument structure are crucial. Using long context models should not be about getting higher agreement, but rather addressing a glaring flaw from a modeling perspective; it is difficult to argue that traditional language models are faithfully modeling aspects of the rubric, such as organization, when essays are being truncated at 512 tokens.

Our modeling results indicate that both the selective attention mechanism and Mamba's linear complexity architecture deliver robust AES performance on lengthy texts. The study's most notable finding is Mamba's exceptional performance despite its simplified architecture. These differences between these models also suggest a potential for ensemble approaches. Several factors position Mamba and related architectures like Jamba \cite{lieber_jamba_2024} as compelling alternatives for large-scale assessment applications. The linear scaling relationship between computational complexity and sequence length offers significant advantages over traditional transformer architectures. Additionally, optimized implementations may achieve 2-8x speed improvements compared to transformer-based models. These efficiency gains, combined with demonstrated effectiveness on long-context tasks, make state space models like Mamba practical solutions for automated assessment and similar applications requiring efficient processing of extended sequences.

\bibliographystyle{acl}
\bibliography{references}

\begin{thebibliography}{41}
\providecommand{\natexlab}[1]{#1}

\bibitem[{{AI@Meta}(2024)}]{aimeta_llama_2024}
{AI@Meta}. 2024.
\newblock \href {https://github.com/meta-llama/llama3/blob/main/MODEL_CARD.md}
  {Llama 3 {Model} {Card}}.

\bibitem[{Beltagy et~al.(2020)Beltagy, Peters, and
  Cohan}]{beltagy_longformer_2020}
Iz~Beltagy, Matthew~E. Peters, and Arman Cohan. 2020.
\newblock \href {https://doi.org/10.48550/arXiv.2004.05150} {Longformer: {The}
  {Long}-{Document} {Transformer}}.
\newblock \emph{arXiv preprint}.
\newblock ArXiv:2004.05150 [cs].

\bibitem[{Bulut et~al.(2024)Bulut, Beiting-Parrish, Casabianca, Slater, Jiao,
  Song, Ormerod, Fabiyi, Ivan, Walsh, Rios, Wilson, Yildirim-Erbasli,
  Wongvorachan, Liu, Tan, and Morilova}]{bulut_rise_2024}
Okan Bulut, Maggie Beiting-Parrish, Jodi~M. Casabianca, Sharon~C. Slater, Hong
  Jiao, Dan Song, Christopher~M. Ormerod, Deborah~Gbemisola Fabiyi, Rodica
  Ivan, Cole Walsh, Oscar Rios, Joshua Wilson, Seyma~N. Yildirim-Erbasli, Tarid
  Wongvorachan, Joyce~Xinle Liu, Bin Tan, and Polina Morilova. 2024.
\newblock \href {https://doi.org/10.48550/arXiv.2406.18900} {The {Rise} of
  {Artificial} {Intelligence} in {Educational} {Measurement}: {Opportunities}
  and {Ethical} {Challenges}}.
\newblock \emph{arXiv preprint}.
\newblock ArXiv:2406.18900.

\bibitem[{Chung et~al.(2022)Chung, Hou, Longpre, Zoph, Tay, Fedus, Li, Wang,
  Dehghani, Brahma, Webson, Gu, Dai, Suzgun, Chen, Chowdhery, Castro-Ros,
  Pellat, Robinson, Valter, Narang, Mishra, Yu, Zhao, Huang, Dai, Yu, Petrov,
  Chi, Dean, Devlin, Roberts, Zhou, Le, and Wei}]{chung_scaling_2022}
Hyung~Won Chung, Le~Hou, Shayne Longpre, Barret Zoph, Yi~Tay, William Fedus,
  Yunxuan Li, Xuezhi Wang, Mostafa Dehghani, Siddhartha Brahma, Albert Webson,
  Shixiang~Shane Gu, Zhuyun Dai, Mirac Suzgun, Xinyun Chen, Aakanksha
  Chowdhery, Alex Castro-Ros, Marie Pellat, Kevin Robinson, and 16 others.
  2022.
\newblock \href {http://arxiv.org/abs/2210.11416} {Scaling
  {Instruction}-{Finetuned} {Language} {Models}}.
\newblock \emph{arXiv preprint}.
\newblock ArXiv:2210.11416 [cs].

\bibitem[{Clark et~al.(2020)Clark, Luong, Le, and Manning}]{clark_electra_2020}
Kevin Clark, Minh-Thang Luong, Quoc~V. Le, and Christopher~D. Manning. 2020.
\newblock \href {https://doi.org/10.48550/arXiv.2003.10555} {{ELECTRA}:
  {Pre}-training {Text} {Encoders} as {Discriminators} {Rather} {Than}
  {Generators}}.
\newblock Technical Report arXiv:2003.10555, arXiv.
\newblock ArXiv:2003.10555 [cs] type: article.

\bibitem[{Crossley et~al.(2022)Crossley, Baffour, Tian, Picou, Benner, and
  Boser}]{crossley_persuasive_2022}
Scott~A. Crossley, Perpetual Baffour, Yu~Tian, Aigner Picou, Meg Benner, and
  Ulrich Boser. 2022.
\newblock \href {https://doi.org/10.1016/j.asw.2022.100667} {The persuasive
  essays for rating, selecting, and understanding argumentative and discourse
  elements ({PERSUADE}) corpus 1.0}.
\newblock \emph{Assessing Writing}, 54:100667.

\bibitem[{Crossley et~al.(2025)Crossley, Baffour, Burleigh, and
  King}]{crossley_large-scale_2025}
Scott~Andrew Crossley, Perpetual Baffour, L.~Burleigh, and Jules King. 2025.
\newblock \href {https://doi.org/10.2139/ssrn.5129353} {A {Large}-{Scale}
  {Corpus} for {Assessing} {Source}-{Based} {Writing} {Quality}: {Asap} 2.0}.

\bibitem[{Dai et~al.(2019)Dai, Yang, Yang, Carbonell, Le, and
  Salakhutdinov}]{dai_transformer-xl_2019}
Zihang Dai, Zhilin Yang, Yiming Yang, Jaime Carbonell, Quoc~V. Le, and Ruslan
  Salakhutdinov. 2019.
\newblock \href {https://doi.org/10.48550/arXiv.1901.02860} {Transformer-{XL}:
  {Attentive} {Language} {Models} {Beyond} a {Fixed}-{Length} {Context}}.
\newblock \emph{arXiv preprint}.
\newblock ArXiv:1901.02860 [cs, stat].

\bibitem[{Dettmers et~al.(2023)Dettmers, Pagnoni, Holtzman, and
  Zettlemoyer}]{dettmers_qlora_2023}
Tim Dettmers, Artidoro Pagnoni, Ari Holtzman, and Luke Zettlemoyer. 2023.
\newblock \href
  {https://proceedings.neurips.cc/paper_files/paper/2023/hash/1feb87871436031bdc0f2beaa62a049b-Abstract-Conference.html}
  {{QLoRA}: {Efficient} {Finetuning} of {Quantized} {LLMs}}.
\newblock \emph{Advances in Neural Information Processing Systems},
  36:10088--10115.

\bibitem[{Devlin et~al.(2018)Devlin, Chang, Lee, and
  Toutanova}]{devlin_bert_2018}
Jacob Devlin, Ming-Wei Chang, Kenton Lee, and Kristina Toutanova. 2018.
\newblock \href {https://doi.org/10.48550/arXiv.1810.04805} {{BERT}:
  {Pre}-training of {Deep} {Bidirectional} {Transformers} for {Language}
  {Understanding}}.
\newblock Technical Report arXiv:1810.04805, arXiv.
\newblock ArXiv:1810.04805 [cs] type: article.

\bibitem[{Dong et~al.(2017)Dong, Zhang, and Yang}]{dong_attention-based_2017}
Fei Dong, Yue Zhang, and Jie Yang. 2017.
\newblock \href {https://doi.org/10.18653/v1/K17-1017} {Attention-based
  {Recurrent} {Convolutional} {Neural} {Network} for {Automatic} {Essay}
  {Scoring}}.
\newblock In \emph{Proceedings of the 21st {Conference} on {Computational}
  {Natural} {Language} {Learning} ({CoNLL} 2017)}, pages 153--162, Vancouver,
  Canada. Association for Computational Linguistics.

\bibitem[{Fu et~al.(2024)Fu, Panda, Niu, Yue, Hajishirzi, Kim, and
  Peng}]{fu_data_2024}
Yao Fu, Rameswar Panda, Xinyao Niu, Xiang Yue, Hannaneh Hajishirzi, Yoon Kim,
  and Hao Peng. 2024.
\newblock \href {https://doi.org/10.48550/arXiv.2402.10171} {Data {Engineering}
  for {Scaling} {Language} {Models} to {128K} {Context}}.
\newblock \emph{arXiv preprint}.
\newblock ArXiv:2402.10171 [cs].

\bibitem[{Graves et~al.(2013)Graves, Jaitly, and Mohamed}]{graves_hybrid_2013}
Alex Graves, Navdeep Jaitly, and Abdel-rahman Mohamed. 2013.
\newblock \href {https://doi.org/10.1109/ASRU.2013.6707742} {Hybrid speech
  recognition with {Deep} {Bidirectional} {LSTM}}.
\newblock In \emph{2013 {IEEE} {Workshop} on {Automatic} {Speech} {Recognition}
  and {Understanding}}, pages 273--278.

\bibitem[{Gu and Dao(2024)}]{gu_mamba_2024}
Albert Gu and Tri Dao. 2024.
\newblock \href {https://doi.org/10.48550/arXiv.2312.00752} {Mamba:
  {Linear}-{Time} {Sequence} {Modeling} with {Selective} {State} {Spaces}}.
\newblock \emph{arXiv preprint}.
\newblock ArXiv:2312.00752 [cs].

\bibitem[{Gu et~al.(2021)Gu, Goel, and Ré}]{gu_efficiently_2021}
Albert Gu, Karan Goel, and Christopher Ré. 2021.
\newblock \href {https://arxiv.org/abs/2111.00396v3} {Efficiently {Modeling}
  {Long} {Sequences} with {Structured} {State} {Spaces}}.

\bibitem[{He et~al.(2021)He, Gao, and Chen}]{he_debertav3_2021}
Pengcheng He, Jianfeng Gao, and Weizhu Chen. 2021.
\newblock \href {https://doi.org/10.48550/arXiv.2111.09543} {{DeBERTaV3}:
  {Improving} {DeBERTa} using {ELECTRA}-{Style} {Pre}-{Training} with
  {Gradient}-{Disentangled} {Embedding} {Sharing}}.
\newblock \emph{arXiv preprint}.
\newblock Number: arXiv:2111.09543 arXiv:2111.09543 [cs].

\bibitem[{Hsieh et~al.(2024)Hsieh, Sun, Kriman, Acharya, Rekesh, Jia, Zhang,
  and Ginsburg}]{hsieh_ruler_2024}
Cheng-Ping Hsieh, Simeng Sun, Samuel Kriman, Shantanu Acharya, Dima Rekesh, Fei
  Jia, Yang Zhang, and Boris Ginsburg. 2024.
\newblock \href {https://doi.org/10.48550/arXiv.2404.06654} {{RULER}: {What}'s
  the {Real} {Context} {Size} of {Your} {Long}-{Context} {Language} {Models}?}
\newblock \emph{arXiv preprint}.
\newblock ArXiv:2404.06654 [cs].

\bibitem[{Hu et~al.(2021)Hu, Shen, Wallis, Allen-Zhu, Li, Wang, Wang, and
  Chen}]{hu_lora_2021}
Edward~J. Hu, Yelong Shen, Phillip Wallis, Zeyuan Allen-Zhu, Yuanzhi Li, Shean
  Wang, Lu~Wang, and Weizhu Chen. 2021.
\newblock \href {https://doi.org/10.48550/arXiv.2106.09685} {{LoRA}:
  {Low}-{Rank} {Adaptation} of {Large} {Language} {Models}}.
\newblock \emph{arXiv preprint}.
\newblock ArXiv:2106.09685 [cs].

\bibitem[{Kaufmann et~al.(2024)Kaufmann, Weng, Bengs, and
  Hüllermeier}]{kaufmann_survey_2024}
Timo Kaufmann, Paul Weng, Viktor Bengs, and Eyke Hüllermeier. 2024.
\newblock \href {https://doi.org/10.48550/arXiv.2312.14925} {A {Survey} of
  {Reinforcement} {Learning} from {Human} {Feedback}}.
\newblock \emph{arXiv preprint}.
\newblock ArXiv:2312.14925.

\bibitem[{Kudo and Richardson(2018)}]{kudo_sentencepiece_2018}
Taku Kudo and John Richardson. 2018.
\newblock \href {https://doi.org/10.18653/v1/D18-2012} {{SentencePiece}: {A}
  simple and language independent subword tokenizer and detokenizer for
  {Neural} {Text} {Processing}}.
\newblock In \emph{Proceedings of the 2018 {Conference} on {Empirical}
  {Methods} in {Natural} {Language} {Processing}: {System} {Demonstrations}},
  pages 66--71, Brussels, Belgium. Association for Computational Linguistics.

\bibitem[{Lieber et~al.(2024)Lieber, Lenz, Bata, Cohen, Osin, Dalmedigos,
  Safahi, Meirom, Belinkov, Shalev-Shwartz, Abend, Alon, Asida, Bergman,
  Glozman, Gokhman, Manevich, Ratner, Rozen, Shwartz, Zusman, and
  Shoham}]{lieber_jamba_2024}
Opher Lieber, Barak Lenz, Hofit Bata, Gal Cohen, Jhonathan Osin, Itay
  Dalmedigos, Erez Safahi, Shaked Meirom, Yonatan Belinkov, Shai
  Shalev-Shwartz, Omri Abend, Raz Alon, Tomer Asida, Amir Bergman, Roman
  Glozman, Michael Gokhman, Avashalom Manevich, Nir Ratner, Noam Rozen, and 3
  others. 2024.
\newblock \href {https://doi.org/10.48550/arXiv.2403.19887} {Jamba: {A}
  {Hybrid} {Transformer}-{Mamba} {Language} {Model}}.
\newblock \emph{arXiv preprint}.
\newblock ArXiv:2403.19887 [cs].

\bibitem[{Loshchilov and Hutter(2019)}]{loshchilov_decoupled_2019}
Ilya Loshchilov and Frank Hutter. 2019.
\newblock \href {https://doi.org/10.48550/arXiv.1711.05101} {Decoupled {Weight}
  {Decay} {Regularization}}.
\newblock \emph{arXiv preprint}.
\newblock ArXiv:1711.05101 [cs, math].

\bibitem[{Lottridge et~al.(2023)Lottridge, Ormerod, and
  Jafari}]{lottridge_psychometric_2023}
Susan Lottridge, Chris Ormerod, and Amir Jafari. 2023.
\newblock Psychometric {Considerations} {When} {Using} {Deep} {Learning} for
  {Automated} {Scoring}.
\newblock In \emph{Advancing {Natural} {Language} {Processing} in {Educational}
  {Assessment}}. Routledge.
\newblock Num Pages: 16.

\bibitem[{McHugh(2012)}]{mchugh_interrater_2012}
Mary~L. McHugh. 2012.
\newblock \href {https://www.ncbi.nlm.nih.gov/pmc/articles/PMC3900052/}
  {Interrater reliability: the kappa statistic}.
\newblock \emph{Biochemia Medica}, 22(3):276--282.

\bibitem[{OpenAI(2023)}]{openai_gpt-4_2023}
OpenAI. 2023.
\newblock \href {https://doi.org/10.48550/arXiv.2303.08774} {{GPT}-4
  {Technical} {Report}}.
\newblock \emph{arXiv preprint}.
\newblock ArXiv:2303.08774 [cs].

\bibitem[{Ormerod et~al.(2023)Ormerod, Burkhardt, Young, and
  Lottridge}]{ormerod_argumentation_2023}
Christopher Ormerod, Amy Burkhardt, Mackenzie Young, and Sue Lottridge. 2023.
\newblock \href {https://doi.org/10.48550/arXiv.2311.06239} {Argumentation
  {Element} {Annotation} {Modeling} using {XLNet}}.
\newblock \emph{arXiv preprint}.
\newblock ArXiv:2311.06239 [cs].

\bibitem[{Ormerod and Kwako(2024)}]{ormerod_automated_2024}
Christopher~Michael Ormerod and Alexander Kwako. 2024.
\newblock \href {https://doi.org/10.48550/arXiv.2407.01873} {Automated {Text}
  {Scoring} in the {Age} of {Generative} {AI} for the {GPU}-poor}.
\newblock \emph{arXiv preprint}.
\newblock ArXiv:2407.01873 [cs].

\bibitem[{Page(2003)}]{page_project_2003}
Ellis~Batten Page. 2003.
\newblock Project {Essay} {Grade}: {PEG}.
\newblock In \emph{Automated essay scoring: {A} cross-disciplinary
  perspective}, pages 43--54. Lawrence Erlbaum Associates Publishers, Mahwah,
  NJ, US.

\bibitem[{Radford et~al.(2018)Radford, Narasimhan, Salimans, and
  Sutskever}]{radford_improving_2018}
Alec Radford, Karthik Narasimhan, Tim Salimans, and Ilya Sutskever. 2018.
\newblock \href
  {https://www.cs.ubc.ca/~amuham01/LING530/papers/radford2018improving.pdf}
  {Improving {Language} {Understanding} by {Generative} {Pre}-training}.

\bibitem[{Rodriguez et~al.(2019)Rodriguez, Jafari, and
  Ormerod}]{rodriguez_language_2019}
Pedro~Uria Rodriguez, Amir Jafari, and Christopher~M. Ormerod. 2019.
\newblock \href {https://doi.org/10.48550/arXiv.1909.09482} {Language models
  and {Automated} {Essay} {Scoring}}.
\newblock \emph{arXiv preprint}.
\newblock Number: arXiv:1909.09482 arXiv:1909.09482 [cs, stat].

\bibitem[{Shermis and Hamner(2013)}]{shermis_contrasting_2013}
Mark~D. Shermis and Ben Hamner. 2013.
\newblock \href {https://doi.org/10.4324/9780203122761.CH19} {Contrasting
  {State}-of-the-{Art} {Automated} {Scoring} of {Essays}}.
\newblock pages 335--368.
\newblock Publisher: Routledge Handbooks Online.

\bibitem[{Su et~al.(2024)Su, Ahmed, Lu, Pan, Bo, and Liu}]{su_roformer_2024}
Jianlin Su, Murtadha Ahmed, Yu~Lu, Shengfeng Pan, Wen Bo, and Yunfeng Liu.
  2024.
\newblock \href {https://doi.org/10.1016/j.neucom.2023.127063} {{RoFormer}:
  {Enhanced} transformer with {Rotary} {Position} {Embedding}}.
\newblock \emph{Neurocomputing}, 568:127063.

\bibitem[{Taghipour and Ng(2016)}]{taghipour_neural_2016}
Kaveh Taghipour and Hwee~Tou Ng. 2016.
\newblock \href {https://doi.org/10.18653/v1/D16-1193} {A {Neural} {Approach}
  to {Automated} {Essay} {Scoring}}.
\newblock In \emph{Proceedings of the 2016 {Conference} on {Empirical}
  {Methods} in {Natural} {Language} {Processing}}, pages 1882--1891, Austin,
  Texas. Association for Computational Linguistics.

\bibitem[{Vaswani et~al.(2017)Vaswani, Shazeer, Parmar, Uszkoreit, Jones,
  Gomez, Kaiser, and Polosukhin}]{vaswani_attention_2017}
Ashish Vaswani, Noam Shazeer, Niki Parmar, Jakob Uszkoreit, Llion Jones,
  Aidan~N Gomez, Łukasz Kaiser, and Illia Polosukhin. 2017.
\newblock \href
  {https://proceedings.neurips.cc/paper/2017/hash/3f5ee243547dee91fbd053c1c4a845aa-Abstract.html}
  {Attention is {All} you {Need}}.
\newblock In \emph{Advances in {Neural} {Information} {Processing} {Systems}},
  volume~30. Curran Associates, Inc.

\bibitem[{Wang et~al.(2019)Wang, Singh, Michael, Hill, Levy, and
  Bowman}]{wang_glue_2019}
Alex Wang, Amanpreet Singh, Julian Michael, Felix Hill, Omer Levy, and
  Samuel~R. Bowman. 2019.
\newblock \href {https://doi.org/10.48550/arXiv.1804.07461} {{GLUE}: {A}
  {Multi}-{Task} {Benchmark} and {Analysis} {Platform} for {Natural} {Language}
  {Understanding}}.
\newblock Technical Report arXiv:1804.07461, arXiv.
\newblock ArXiv:1804.07461 [cs] type: article.

\bibitem[{Warner et~al.(2024)Warner, Chaffin, Clavié, Weller, Hallström,
  Taghadouini, Gallagher, Biswas, Ladhak, Aarsen, Cooper, Adams, Howard, and
  Poli}]{warner_smarter_2024}
Benjamin Warner, Antoine Chaffin, Benjamin Clavié, Orion Weller, Oskar
  Hallström, Said Taghadouini, Alexis Gallagher, Raja Biswas, Faisal Ladhak,
  Tom Aarsen, Nathan Cooper, Griffin Adams, Jeremy Howard, and Iacopo Poli.
  2024.
\newblock \href {https://doi.org/10.48550/arXiv.2412.13663} {Smarter, {Better},
  {Faster}, {Longer}: {A} {Modern} {Bidirectional} {Encoder} for {Fast},
  {Memory} {Efficient}, and {Long} {Context} {Finetuning} and {Inference}}.
\newblock \emph{arXiv preprint}.
\newblock ArXiv:2412.13663 [cs].

\bibitem[{Williamson et~al.(2012)Williamson, Xi, and
  Breyer}]{williamson_framework_2012}
David~M. Williamson, Xiaoming Xi, and F.~Jay Breyer. 2012.
\newblock \href {https://doi.org/10.1111/j.1745-3992.2011.00223.x} {A
  {Framework} for {Evaluation} and {Use} of {Automated} {Scoring}}.
\newblock \emph{Educational Measurement: Issues and Practice}, 31(1):2--13.
\newblock \_eprint:
  https://onlinelibrary.wiley.com/doi/pdf/10.1111/j.1745-3992.2011.00223.x.

\bibitem[{Wu et~al.(2019)Wu, Fan, Baevski, Dauphin, and Auli}]{wu_pay_2019}
Felix Wu, Angela Fan, Alexei Baevski, Yann~N. Dauphin, and Michael Auli. 2019.
\newblock \href {https://doi.org/10.48550/arXiv.1901.10430} {Pay {Less}
  {Attention} with {Lightweight} and {Dynamic} {Convolutions}}.
\newblock \emph{arXiv preprint}.
\newblock ArXiv:1901.10430 [cs].

\bibitem[{Xiao et~al.(2024)Xiao, Ma, Xu, Zhang, Wang, and
  Fu}]{xiao_automation_2024}
Changrong Xiao, Wenxing Ma, Sean~Xin Xu, Kunpeng Zhang, Yufang Wang, and Qi~Fu.
  2024.
\newblock \href {http://arxiv.org/abs/2401.06431} {From {Automation} to
  {Augmentation}: {Large} {Language} {Models} {Elevating} {Essay} {Scoring}
  {Landscape}}.
\newblock \emph{arXiv preprint}.
\newblock ArXiv:2401.06431 [cs] version: 1.

\bibitem[{Xu et~al.(2023)Xu, Xie, Qin, Tao, and
  Wang}]{xu_parameter-efficient_2023}
Lingling Xu, Haoran Xie, Si-Zhao~Joe Qin, Xiaohui Tao, and Fu~Lee Wang. 2023.
\newblock \href {https://doi.org/10.48550/arXiv.2312.12148}
  {Parameter-{Efficient} {Fine}-{Tuning} {Methods} for {Pretrained} {Language}
  {Models}: {A} {Critical} {Review} and {Assessment}}.
\newblock \emph{arXiv preprint}.
\newblock ArXiv:2312.12148 [cs].

\bibitem[{Yang et~al.(2019)Yang, Dai, Yang, Carbonell, Salakhutdinov, and
  Le}]{yang_xlnet_2019}
Zhilin Yang, Zihang Dai, Yiming Yang, Jaime Carbonell, Russ~R Salakhutdinov,
  and Quoc~V Le. 2019.
\newblock \href
  {https://proceedings.neurips.cc/paper/2019/hash/dc6a7e655d7e5840e66733e9ee67cc69-Abstract.html}
  {{XLNet}: {Generalized} {Autoregressive} {Pretraining} for {Language}
  {Understanding}}.
\newblock In \emph{Advances in {Neural} {Information} {Processing} {Systems}},
  volume~32. Curran Associates, Inc.

\end{thebibliography}

\end{document}